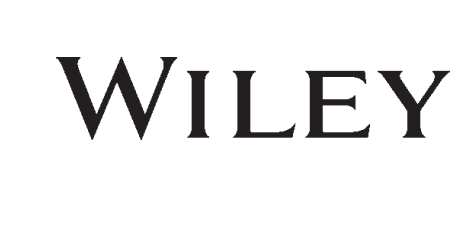

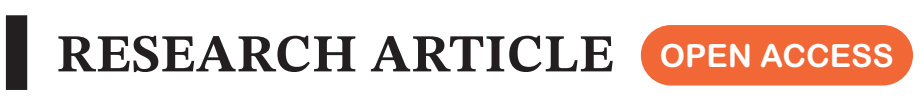

RESEARCH ARTICLE OPEN ACCESS

# Transforming Hyperspectral Images Into Chemical Maps: A Novel End-to-End Deep Learning Approach

Ole-Christian Galbo Engstrøm[1,2,3] | Michela Albano-Gaglio[4] | Erik Schou Dreier[1] | Yamine Bouzembrak[5] | Maria Font-i-Furnols[4] | Puneet Mishra[6] | Kim Steenstrup Pedersen[2,7]

[1]FOSS Analytical A/S, Hillerød, Denmark | [2]Department of Computer Science (DIKU), University of Copenhagen, Copenhagen, Denmark | [3]Department of Food Science (UCPH FOOD), University of Copenhagen, Frederiksberg, Denmark | [4]IRTA - Food Quality and Technology, Finca Camps i Armet, Girona, Spain | [5]Information Technology Group, Wageningen University and Research (WUR), Wageningen, the Netherlands | [6]Food and Biobased Research, Wageningen University and Research (WUR), Wageningen, the Netherlands | [7]Natural History Museum of Denmark (NHMD), University of Copenhagen, Copenhagen, Denmark

**Correspondence:** Puneet Mishra (puneet.mishra@wur.nl)

**Received:** 18 April 2025 | **Revised:** 14 June 2025 | **Accepted:** 17 June 2025

**Funding:** This work was supported by The Innovation Fund Denmark and FOSS Analytical A/S (grant number 1044-00108B); FEDER and MICIU/AEI/10.13039/501100011033/ (grant number RTI2018-096993-B-I00, 2019–2022); and the Spanish National Institute of Agricultural Research (INIA) (grant number PRE2019-089669, 2020–2024).

## ABSTRACT

Current approaches to chemical map generation from hyperspectral images are based on models such as partial least squares (PLS) regression, generating pixel-wise predictions that do not consider spatial context and suffer from a high degree of noise. This study proposes an end-to-end deep learning approach using a modified version of U-Net and a custom loss function to directly obtain chemical maps from hyperspectral images, skipping all intermediate steps required for traditional pixel-wise analysis. The U-Net is compared with the traditional PLS regression on a real dataset of pork belly samples with associated mean fat reference values. The U-Net obtains a test set root mean squared error that is 7% lower than that of PLS regression on the task of mean fat prediction. At the same time, U-Net generates fine detail chemical maps where 99.91% of the variance is spatially correlated. Conversely, only 2.37% of the variance in the PLS-generated chemical maps is spatially correlated, indicating that each pixel-wise prediction is largely independent of neighboring pixels. Additionally, while the PLS-generated chemical maps contain predictions far beyond the physically possible range of 0%–100%, U-Net learns to stay inside this range. Thus, the findings of this study indicate that U-Net is superior to PLS for chemical map generation.

## 1 | Introduction

Hyperspectral imaging (HSI) is an advanced analytical technique that enables the acquisition of spatially distributed spectral information from material surfaces [1]. Unlike traditional spectroscopy, which provides only bulk spectral data, HSI simultaneously captures spectral and spatial information, offering a more comprehensive analysis of heterogeneous samples.

HSI is widely applied across various scientific disciplines and implemented in multiple spectral modalities, including near-infrared (NIR) [2], Raman [3], and infrared (IR) [4] spectroscopy. The fundamental distinction between HSI and conventional spectroscopy lies in its imaging capability, where each pixel of an image contains a complete spectral profile. This enables the generation of chemical maps representing the spatial distribution of specific chemical components within a sample. However, due to the high-dimensional nature

[Correction added on 17 November 2025, after first online publication: This work is an update of the original, previously published work, and is the preferred version.]

of HSI data, advanced data modeling techniques are required to extract meaningful and interpretable information. The ability to visualize compositional variations in an image format makes HSI a powerful tool in diverse applications, including material science, pharmaceuticals, food quality assessment, and biomedical imaging.

Chemometric modeling plays a crucial role in HSI data processing [1, 5], particularly in the generation of chemical maps. Traditionally, chemical maps are generated using supervised or unsupervised methods, where each pixel is analyzed independently. For instance, supervised models are typically trained using mean spectra and subsequently applied to predict chemical properties for individual pixels [5]. In the case of supervised modeling, it is essential to note that the minimum sample size required for chemical analysis is often larger than the pixel size in hyperspectral images. Consequently, acquiring pixel-wise reference values for chemical maps is generally infeasible. Instead, reference values are obtained as averaged measurements at a much lower resolution, such as over an entire sample or larger subsets thereof.

Because reference values represent sample-wise means, it is common practice to train models to predict these mean values. A straightforward approach involves averaging the NIR hyperspectral image pixels to generate a mean spectrum, which is then paired with its corresponding mean reference value. These data pairs are typically used to train a Partial Least Squares (PLS) regression model [6, 7], which is subsequently applied to generate pixel-wise predictions, producing chemical maps [8–11]. However, this approach has several limitations. Traditional PLS-based chemical maps treat each pixel independently, resulting in predictions with a low degree of spatial correlation. Pixel-wise predictions may fall outside physically meaningful ranges (0%–100%), affecting interpretability. The lack of spatial structure limits the ability to assess variations within samples, thereby constraining downstream optimization.

The significance of spatial and contextual information in HSI data modeling is now widely recognized in the chemometric domain [12, 13]. Recent studies have demonstrated the benefits of integrating spatial information alongside spectral data to enhance model accuracy [14–16]. These approaches offer improved pixel-level property prediction; however, they do not fundamentally enhance spatial coherence because the chemical maps are still generated in a pixel-wise manner.

This study aims to bridge the gap between spectral-based models, such as PLS, and the inherently spatio-spectral task of chemical map generation. To this end, a novel method for chemical map generation is proposed. A modification of U-Net [17], a convolutional neural network (CNN) originally designed for semantic segmentation in medical imaging, is proposed for chemical map generation. In semantic segmentation, each pixel in an input image is to be classified into discrete categories. The equivalent task for chemical map generation involves continuous regression rather than classification. There are several advantages of using U-Net for chemical mapping. Unlike PLS, which processes pixels independently, U-Net jointly considers spatial and spectral information within hyperspectral images. It performs exceptionally well with limited labeled samples [18–21]. Like PLS models trained on mean spectra, the proposed U-Net requires only mean reference values per sample. A custom multi-objective loss function trains U-Net to generate pixel-wise predictions strictly inside the physically valid range (0%–100%) and leverages spatial structure to generate smoothly varying chemical maps, neither of which PLS accomplishes.

This study proposes a new deep learning approach (a modified U-Net) to process HSI data. The study utilizes the NIR-HSI dataset of pork bellies and associated mean fat content reference values from Albano-Gaglio et al. [8] to compare PLS-based and U-Net–based chemical map generation approaches. The remainder of this article is structured as follows: Section 2 presents details on the dataset and modeling approaches. Section 3 and Section 4 analyze and discuss the experimental results. Section 5 provides conclusions on the study and future perspectives on using deep learning for chemical mapping.

## 2 | Materials and Methods

### 2.1 | Dataset

This study used the VIS-NIR reflectance HSI dataset of pork bellies introduced by Albano-Gaglio et al. [8]. Bellies were selected from different production systems, feedings, genetics, and sexes to ensure enough variability in their quality parameters to calibrate non-destructive devices and methods, such as HSI systems, to characterize pork bellies. Due to the selection made by Albano-Gaglio et al. [8], the pork bellies belong to pigs spanning five different fat classes, F1–F5, with F1–F3 from common commercial pigs, F4 from Duroc pigs, and F5 from Iberian crossbred pigs. The dataset consists of hyperspectral images of 182 pork bellies. Figure 1 illustrates an entire pork belly. To simplify sampling, each pork belly was divided into five vertical sections, 1–5, and three horizontal sections, A-C. Then, Albano-Gaglio et al. [8] physically cut each pork belly to yield five slices, corresponding to the five vertical sections.

The division of 182 pork bellies into five slices each would give 910 slices in total, but as nine slices are missing from the dataset, we end up with 901. The 901 slices represent all 182 pork bellies. Albano-Gaglio et al. [8] imaged each slice using a VIS-NIR platform with 300 uniformly distributed wavelength channels from 386.63 nm to 1015.78 nm. All the images have a width of 900 pixels, while their height varies between 1900 and 2477 pixels. Most images have a height of 2000 pixels, which leads to a mean height of 1982 pixels. Only the latter 124 wavelength channels from 747.92 nm upwards were used for modeling. Using this range serves two purposes. First, it reduces the number of bands, which reduces the computation cost for model training. Second, it allows the models to focus only on the chemical information to predict fat content rather than biasing them with the color information.

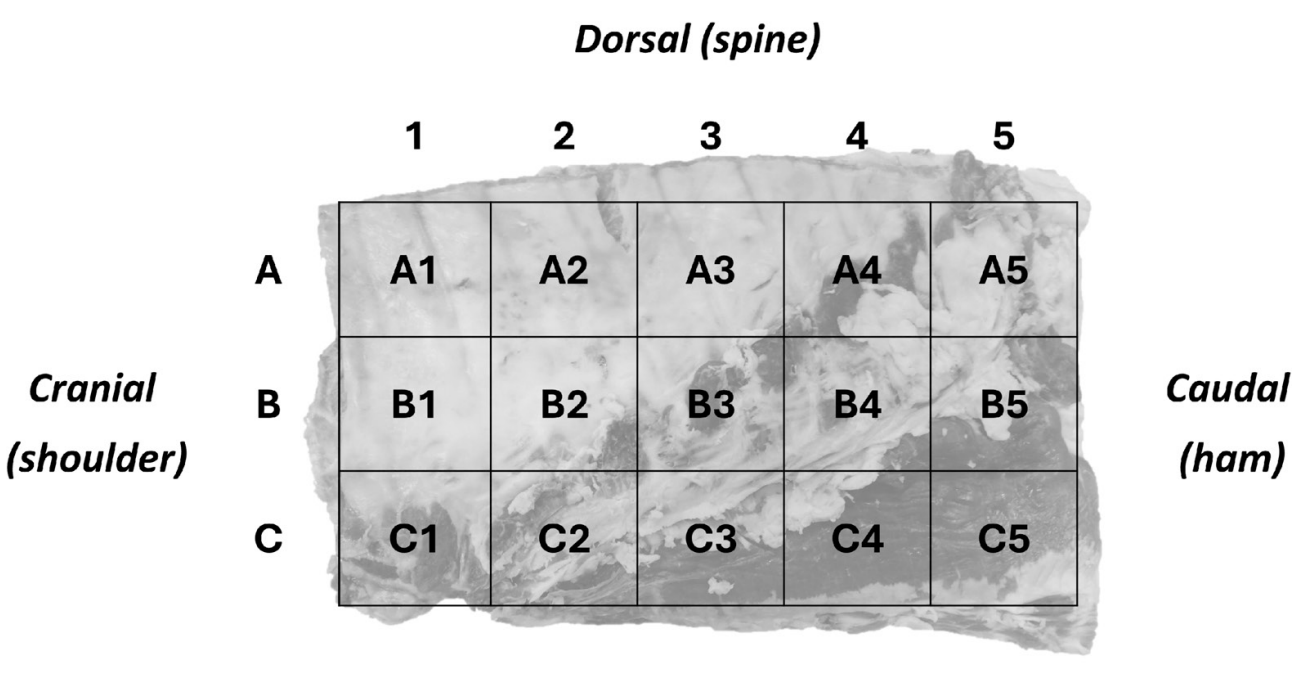

**FIGURE 1** | An entire pork belly. Illustration kindly donated by Albano-Gaglio et al. [8]. Each belly was physically cut into five pieces, corresponding to sections 1–5, before being imaged.

Albano-Gaglio et al. [8] associate each image with a segmentation mask computed using PLS discriminant analysis (PLS-DA, [22]) to distinguish between the slice and the background. The segmentation mask also classifies muscle tissue and other stains as background, retaining only the subcutaneous fat tissue. In this study, the masks were slightly modified by performing a morphological erosion with a disk-shaped structuring element with a radius of one pixel. This erosion ensures that all pixels in the 4-neighborhood of every retained foreground pixel in the hyperspectral image are known to contain only foreground initially. This, in turn, means that the spatial derivatives in both directions of every retained foreground pixel are well-defined, a property required when computing these spatial derivatives in Section 3 to quantify the spatial distribution of predicted chemical maps. An example image and eroded mask is shown in Figure 2. In practice, due to the small size of the structuring element, the eroded mask is practically identical to the original mask. From now on, unless explicitly stated otherwise, when referring to the image's mask, the eroded version is meant. With each pork belly, a mean spectrum was computed as the mean of all spectra contained within the masks of the images of vertical slices belonging to that belly. As a chemical parameter, this study focuses on the fat content of the whole minced belly, the reference value of which Albano-Gaglio et al. [8] obtained with wet chemistry.

#### 2.1.1 | Dataset Split

Albano-Gaglio et al. [8] split the dataset into two subsets, cross-validation (CV) and test, with the DUPLEX algorithm [23]. The DUPLEX algorithm generates two datasets in an iterative process based on Mahalanobis distance [24] between the mean spectra of the pork bellies.

In the dataset split provided by Albano-Gaglio et al. [8], the CV set consists of 122 pork bellies, while the test set consists of 60 bellies. Albano-Gaglio et al. [8] randomly split the CV set into five folds (subsets). To make the results of this study comparable to those of Albano-Gaglio et al. [8], the same splitting of the CV set into folds from Albano-Gaglio et al. [8] was requested. However, as the randomly generated splits were not stored, this request could not be fulfilled. Instead, a modification of the DUPLEX algorithm to generate not two but six subsets of the dataset was explored. The original test set was kept as one of the subsets to enable a direct comparison between the test set results and those of Albano-Gaglio et al. [8]. The modified DUPLEX algorithm divides the CV set into five folds. Before feeding the mean spectra to DUPLEX, a dimensionality reduction using principal component analysis (PCA, [25, 26]) was performed using six components, which was the minimum required to retain at least 99% of the variance of the CV subset. In Table 1, a description of the number of pork bellies and how associated slices are distributed between the dataset splits yielded by DUPLEX is presented. In Figure 3, distributions of reference fat percentages of these dataset splits are presented. All fat classes, F1–F5, are represented in the CV and test sets.

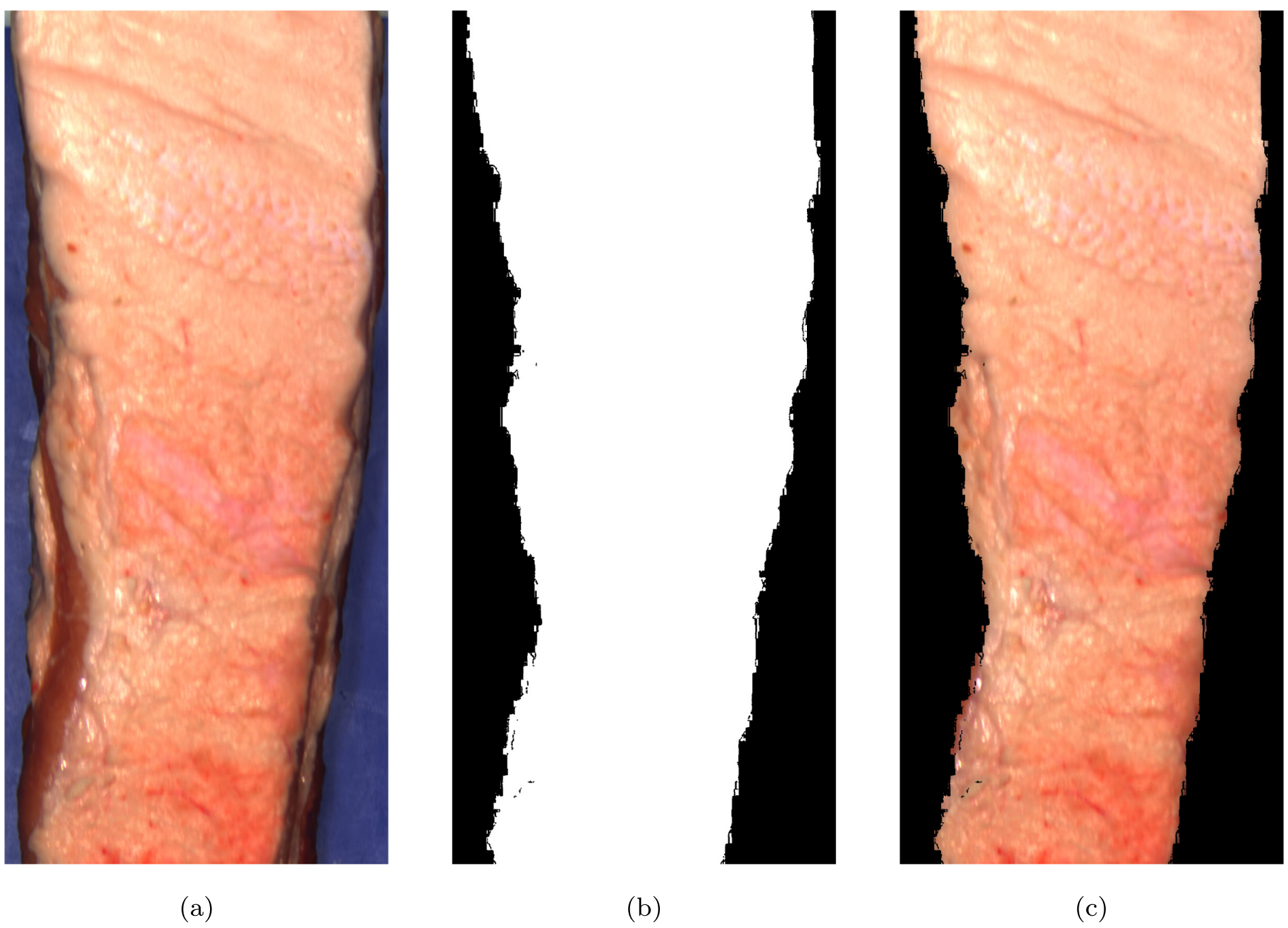

**FIGURE 2** | RGB illustrations of an example image and associated eroded mask from the dataset. This image corresponds to one of the five vertical sections of a pork belly, such as illustrated in Figure 1. (a) Image of pork belly slice. (b) Associated segmentation mask eroded by a disk-shaped structuring element with radius one. (c) Masked image of pork belly slice.

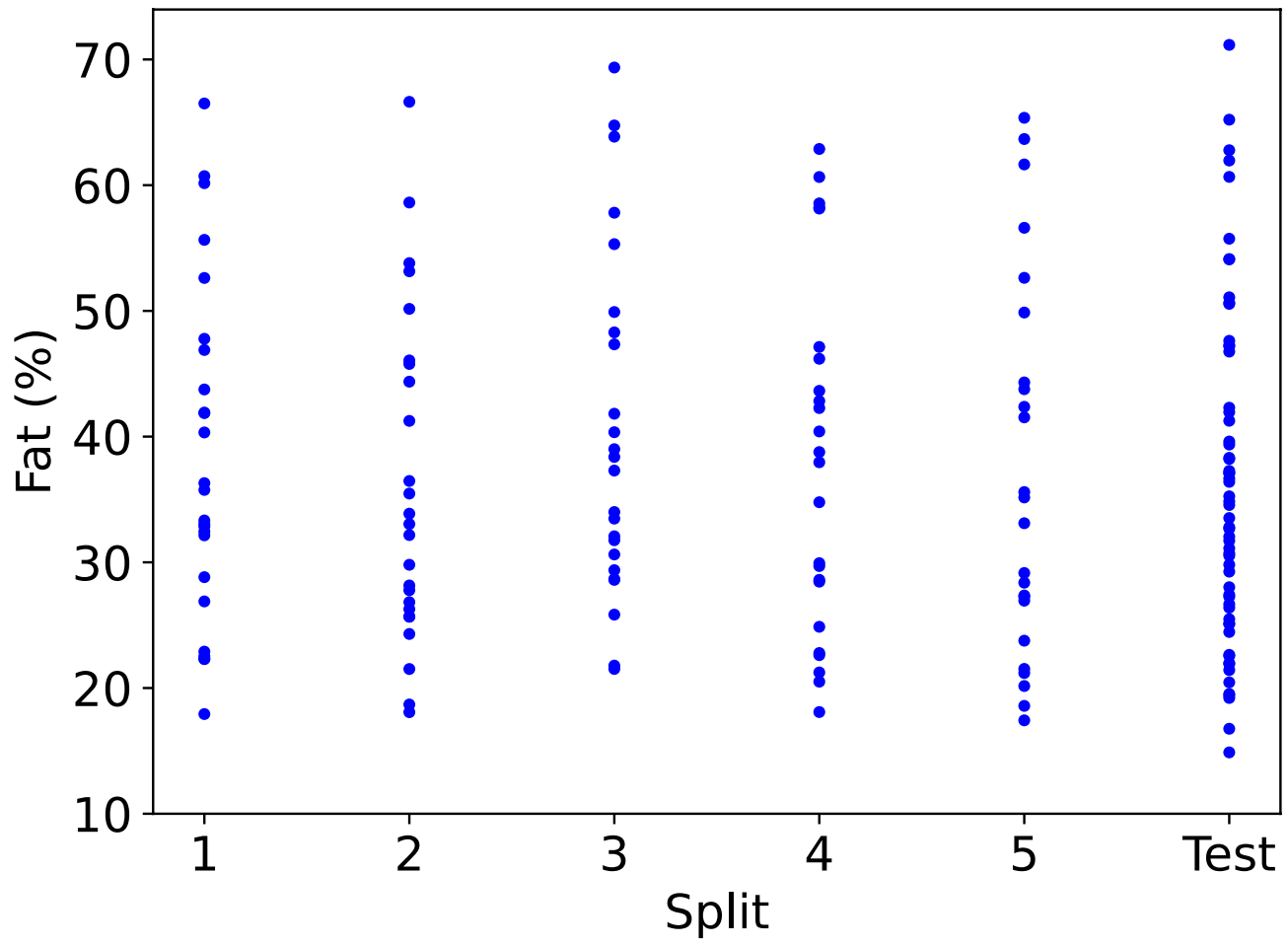

**FIGURE 3** | Distribution of reference fat values in the dataset splits.

**TABLE 1** | Number of pork bellies and slices of those bellies for each dataset split.

| | Split | | | | | | |
|---|---|---|---|---|---|---|---|
| | **1** | **2** | **3** | **4** | **5** | **Test** | **Total** |
| No. of bellies | 25 | 25 | 24 | 24 | 24 | 60 | 182 |
| No. of slices | 123 | 125 | 119 | 118 | 118 | 298 | 901 |

*Note:* We have reference values for the bellies and images of the slices.

## 2.2 | Modeling

### 2.2.1 | PLS

A PLS model using the pork bellies' mean absorbance spectra was calibrated as a baseline. For a pork belly, the mean absorbance spectrum was calculated by first taking the negative logarithm of each pixel in each hyperspectral image of each pork belly slice and subsequently preprocessing the pixels by application of Standard Normal Variate (SNV) transformation [27] followed by convolution with a Savitzky-Golay filter [28, 29] with a window length of seven, a polynomial order of two, and a derivative order of two. Subsequently, a mean spectrum for each pork belly was computed by averaging all pixels inside the slices' masks. These preprocessing steps combat Rayleigh scattering and reveal the chemical information in the spectra [30, 31]. Finally, a column-wise mean centering of the spectra and the reference fat values was performed using statistics computed only on the training sets.

A five-fold cross-validation was used to determine the optimal number of components. For each CV fold, the root mean squared error (RMSE) on the validation set was computed for all possible numbers of components. Subsequently, the average over the five folds was calculated to get a mean RMSE for each number of components. This evaluation yielded 11 components minimizing the RMSE, so a PLS on the entire CV set using 11 components was recalibrated. Thus, the final PLS model was calibrated using all the available data points in the CV set. The PLS calibrations were carried out using the fast [32] and numerically stable [33] Improved Kernel PLS Algorithm 1 [34] with the software implemented by Engstrøm et al. [35]. The CV was done with the same software using the fast algorithm by Engstrøm and Jensen [36].

In addition to predicting from the mean spectra for each pork belly, this same PLS model was used to generate chemical maps by pixel-wise prediction as traditionally done in HSI processing [1, 5]. The same preprocessing was applied to each pixel as for the mean spectra described above. Thus, PLS predictions on mean spectra are equivalent to the average of pixel-wise predictions across chemical maps of slices of the same pork bellies.

### 2.2.2 | U-Net

U-Net was modified in two ways: The first modification was replacing the upsampling by transposed convolution, as proposed by Ronneberger et al. [17], with upsampling by bilinear interpolation. Transposed convolutions cause square-like artifacts in the output [37]. Bilinear interpolation avoids these artifacts while reducing the number of parameters in the model, thereby decreasing training time.

The second modification was adding an initial 3D convolution layer with one filter containing one kernel of size (depth, height, width) = (7, 2, 2). Here, the three-dimensional filter was applied across the three spatio-spectral dimensions. An initial 3D convolution layer before a 2D neural network allows learning spectral smoothing and derivatives. Choosing a depth of seven enables the model to learn precisely the Savitzky-Golay filter used to preprocess spectra for PLS. Adding this layer is similar to the approach taken by Engstrøm et al. [38, 39], where it was shown to be beneficial for predicting chemical parameters in NIR-HSI images of grain without being detrimental to non-chemical parameters.

Appendix A illustrates the model architecture and the sizes of intermediate tensors. Apart from these modifications, the new design followed the original U-Net implementation. As in the original U-Net implementation, no padding to any convolutions was applied, and only the so-called valid parts of the convolutions were evaluated. The valid parts of convolution are those pixels where the kernel completely overlaps with the input image or intermediate feature map. All convolution layers had their bias initialized to zero and their scale parameters initialized by sampling from a Kaiming He Normal Distribution [40]. While Ronneberger et al. [17] wrote the original U-Net implementation in Caffe [41], this study recreated it in PyTorch [42] to accommodate the customization needs better and to ensure easy compatibility and integration with other modern works.

For the initial 3D convolution layer, a stride of 2 pixels was applied for the spatial dimensions while keeping a stride of 1 channel for the depth dimension. The stride effectively halves the input size from a height and width of $2360 \times 1272$ to $1180 \times 636$ and serves to significantly reduce training time and reduce the size of the output. Feeding this input to the U-Net gives an output of size $996 \times 452$. These dimensions align so that U-Net does not drop any rows or columns of its intermediate feature maps in any layers, even with max-pooling layers with a stride of two pixels and non-padded convolution layers. Using these fixed dimensions requires data augmentation as the original dataset contains images with a fixed width of 900 pixels and a varying height.

Similarly, the size of the spectral dimension was reduced by a factor of two from 124 channels to 62 channels. However, instead of strided convolutions, the spectral dimension was reduced by averaging neighboring wavelength channels. This reduction simultaneously reduces training time and alleviates the so-called curse of dimensionality, which is prevalent for deep learning on hyperspectral images [43].

In the original design, U-Net produces an output smaller than its input. This shrinking implies that padding must be applied to retain the size of the original input. This study used a two-stage scheme for padding the images before feeding them to the U-Net model. The first stage applies background padding to the height and width until reaching 1992 pixels in height and 904 pixels in width. The background spectrum used for padding is the average spectrum of the image's left-most and right-most columns, as these contain almost exclusively background for all images. The justification for stage one is that, with this study's chosen dimensions, the model will produce an output corresponding to the central 1992 pixels in height and 904 pixels in width. Thus, if an image was smaller than this size, it was padded with a background spectrum. Due to using only the valid parts of the convolution, the input image must be further padded to get predictions near the border. In this second stage, an approach similar to the original U-Net design was taken, and, as such, mirror-padding was applied until reaching 2360 pixels in height and 1272 pixels in width. Illustrations of the padding scheme are shown in Figure 4.

The mask must be modified according to the padding scheme to match the contents of the central $1992 \times 904$ pixels of the input, padded hyperspectral image. The original non-eroded mask was used as the starting point for these modifications. Any location padded with the background spectrum was accordingly set to background in the mask. Then, the central $1992 \times 904$ pixels of the mask were cropped as these correspond to the pixels for which U-Net makes a prediction. Due to the initial strided 3D convolution, the spatial dimensions of the input to U-Net are halved. Accordingly, the spatial dimensions of the mask must be halved to $996 \times 452$, matching the output of the U-Net. The halving of the mask is computed with bilinear interpolation, after which every pixel in the mask is rounded to zero or one. Finally, the downsampled mask is eroded with a disk-shaped structuring element with a radius of one pixel to allow computation of spatial derivatives in the predictions as explained in Section 2.1. In the context of U-Net, a reference to an image's mask means this downsampled, eroded version.

During training, images were randomly flipped horizontally and vertically, each with a probability of 0.5. Such flipping was not applied during evaluation.

The task to be solved must be defined to optimize the modified U-Net model. Recall that the data contains images of slices of pork belly and reference values for the mean fat percentage of the whole bellies. This reference value was used for each slice for optimization, an approach similar to Engstrøm et al. [38, 39]. Therefore, given an input image of a slice, the model aims to predict a fat distribution with a mean equal to the fat reference. Only the pixels inside the mask were considered for a given predicted chemical map. Theoretically, a model can be trained to predict a fat value of zero outside the mask. However, as seen in Figure 2, the edges of the meat are often classified as background by the mask. Thus, in practice, attempting to learn to predict zero outside the mask with this dataset will likely be detrimental to the overall performance of the model.

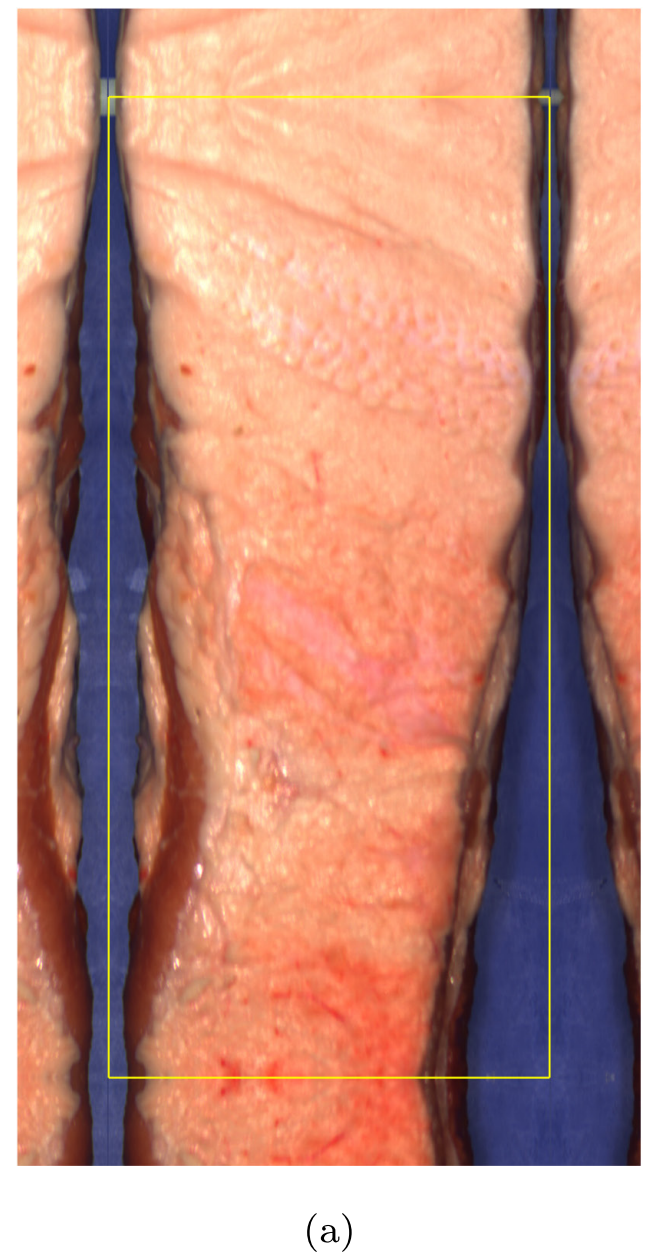

(a)

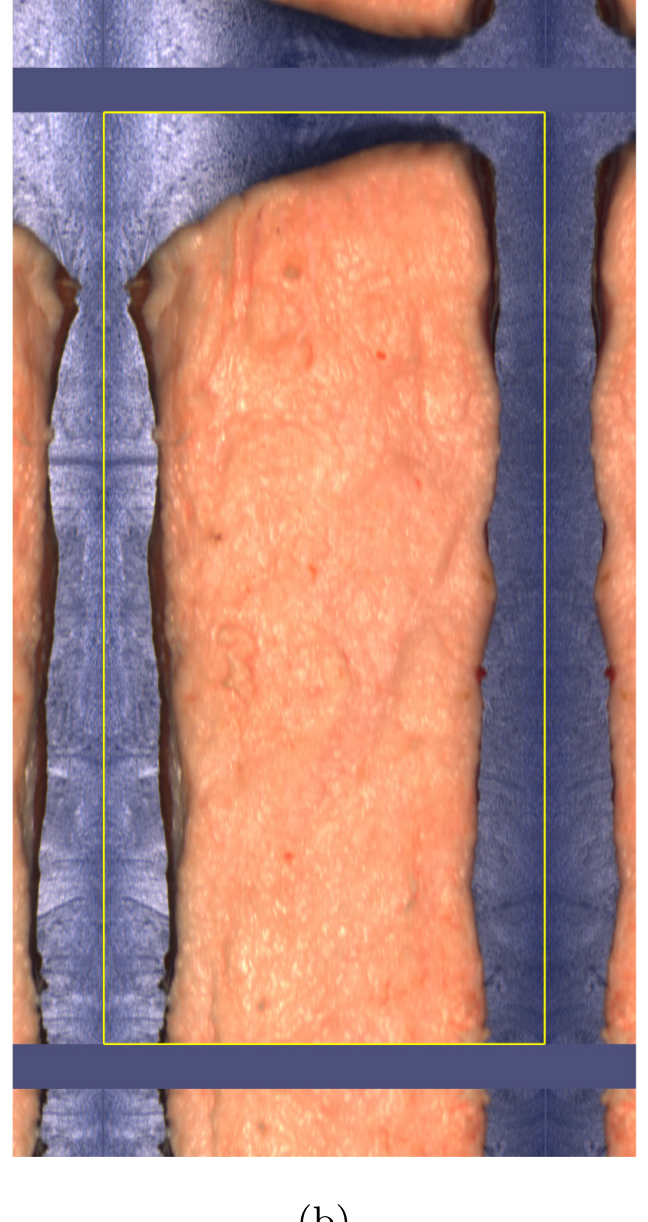

(b)

**FIGURE 4** | Example applications of the two-stage padding scheme. The yellow boxes indicate the border between the original images and the padding. (a) Padding of an image with an original height of at least 1992 pixels. Here, two columns of background padding are applied to the left and right, reaching a width of 904 pixels. Then, mirror-padding is applied until a height of 2360 and a width of 1272 pixels. This is the padded version of the image shown in Figure 2a. (b) Padding of an image with an original height of less than 1992 pixels. Here, background padding is applied until reaching a height of 1992 and a width of 904 pixels, after which mirror-padding is applied until reaching a height of 2360 and a width of 1272 pixels.

Given a batch of hyperspectral images, let $\hat{\mathbf{Y}} \in \mathbb{R}^{B \times H \times W}$ be the U-Net's predicted chemical maps with batch size $B$, height $H$ pixels, and width $W$ pixels. For a given batch, consider also the associated reference fat values $\mathbf{y} \in \mathbb{R}^{B}$ and segmentation masks $\mathbf{M} \in \{0,1\}^{B \times H \times W}$. In $\mathbf{M}$, a value of 0 indicates background and a value of 1 indicates foreground in the corresponding pixel of $\hat{\mathbf{Y}}$. Such pixels are considered outside, respectively, inside the mask. For a batch of images, consider the modified U-Net's masked prediction, $\hat{\mathbf{Y}}^{\mathbf{M}}$, defined using element-wise multiplication ($\odot$)

$$\hat{\mathbf{Y}}^{\mathbf{M}} = \hat{\mathbf{Y}} \odot \mathbf{M} \tag{1}$$

Then, the U-Net's mean fat predictions, $\hat{\mathbf{y}} \in \mathbb{R}^{B}$, are defined as its mean predictions inside the masks. Considering $\hat{\mathbf{y}}$ at batch index $b \in \{1, \ldots, B\}$, the mean fat prediction is given by

$$\hat{\mathbf{y}}_b = \frac{\sum_{h=1}^{H} \sum_{w=1}^{W} \hat{\mathbf{Y}}^{\mathbf{M}}_{b,h,w}}{\sum_{h=1}^{H} \sum_{w=1}^{W} \mathbf{M}_{b,h,w}} \tag{2}$$

Now, the modified U-Net can be optimized by punishing it for predicting an average value that deviates from the reference value. To this end, the mean squared error between the reference value $y_b$ and $\hat{y}_b$ was used

$$MSE(\mathbf{y}, \hat{\mathbf{y}}) = \frac{1}{B} \sum_{b=1}^{B} \left(\mathbf{y}_b - \hat{\mathbf{y}}_b\right)^2 \tag{3}$$

The true fat contents in each pixel must lie in the range 0%–100%. However, only optimizing against the mean fat prediction provides no guarantee for a proper range of pixel-wise predictions. Thus, to aid the optimization process, the out-of-bounds loss (OOBL) was defined as the sum of squared deviations from the 0–100 range in $\hat{\mathbf{Y}}^{\mathbf{M}}$

$$OOBL\left(\hat{\mathbf{Y}}^{\mathbf{M}}\right) = \frac{1}{B} \sum_{b=1}^{B} \sum_{h=1}^{H} \sum_{w=1}^{W} \max\left(-\hat{\mathbf{Y}}^{\mathbf{M}}_{b,h,w}, 0\right)^2 + \max\left(\hat{\mathbf{Y}}^{\mathbf{M}}_{b,h,w} - 100, 0\right)^2. \tag{4}$$

Here, the first term in the sum considers pixel-wise predictions below 0, and the latter considers those above 100. No normalization by the sizes of the masks was applied because an out-of-bounds prediction is considered equally nonsensical, regardless of the size of the pork belly slice on which the prediction lies.

In addition to having the correct mean fat value and a pixel-wise range of 0%–100%, a chemical map is expected to have spatial dependencies among the pixels. In particular, it was assumed that a pixel and its neighbors should have predictions that are not too far apart, which means that the pixel values of the predicted chemical map should be smoothly varying. Therefore, a loss term was added to punish the model for deviations from this assumption. It computes a measure of the predicted chemical map's total smoothness and is referred to as the smoothness loss (SL). The sum of the squared spatial gradient magnitude of the predicted chemical map was used as a measure of smoothness, and SL is defined as the average over the batch

$$SL\left(\hat{\mathbf{Y}}, \mathbf{M}\right) = \frac{1}{B} \sum_{b=1}^{B} \frac{\sum_{h=1}^{H-1} \sum_{w=1}^{W-1} \mathbf{M}_{b,h,w} \left( \left(\hat{\mathbf{Y}}_{b,h,w} - \hat{\mathbf{Y}}_{b,h+1,w}\right)^2 + \left(\hat{\mathbf{Y}}_{b,h,w} - \hat{\mathbf{Y}}_{b,h,w+1}\right)^2 \right)}{\sum_{h=1}^{H-1} \sum_{w=1}^{W-1} \mathbf{M}_{b,h,w}} \tag{5}$$

The squared spatial gradient magnitude is approximated by approximating the spatial first-order derivatives with forward differences between horizontal and vertical neighboring pixel pairs in the predicted chemical map, considering only pairs where both pixels are meat. Consider a pixel, $\hat{\mathbf{Y}}_{b,h,w}$. If it lies inside the mask, then, due to the previous erosion of the mask, the neighbor pixels $\hat{\mathbf{Y}}_{b,h+1,w}$ and $\hat{\mathbf{Y}}_{b,h,w+1}$ are all known to contain meat regardless of whether they lie inside the mask or not. This property is necessary and sufficient for the local derivative of the predicted chemical map at $\hat{\mathbf{Y}}_{b,h,w}$ to be well-defined over the meat because the derivative is then not affected by the background. Enforcing this constraint is sensible, as the quality of the predicted chemical map is only related to the smoothness inside the chemical map and not to the smoothness between the chemical map and the background. SL is normalized by the mask size to get an average smoothness measure for each predicted chemical map.

L2-regularization was also added to the loss function to combat potential overfitting. The modified U-Net's weights were organized into a vector, and that vector's Euclidean (L2) distance[1] was computed. Bias parameters were not regularized. Let $\boldsymbol{\theta}$ be the modified U-Net's weights, indexed by $p$ such that $\boldsymbol{\theta}_p$ is the $p$'th weight. Then, considering a total of $P$ weights, the L2-regularization loss is given by

$$L2(\boldsymbol{\theta}) = \sum_{p=1}^{P} \boldsymbol{\theta}_p^2 \tag{6}$$

Finally, given $\boldsymbol{\theta}$, $\hat{\mathbf{Y}}$, $\mathbf{M}$, and $\mathbf{y}$, Equation (1) was used to compute $\hat{\mathbf{Y}}^{\mathbf{M}}$ and Equation (2) was used to compute $\hat{\mathbf{y}}$, allowing to express the total loss (TL) as

$$\begin{aligned} TL\left(\boldsymbol{\theta}, \hat{\mathbf{Y}}, \mathbf{M}, \mathbf{y}, \hat{\mathbf{Y}}^{\mathbf{M}}, \hat{\mathbf{y}}\right) = {} & \lambda_{MSE} MSE(\mathbf{y}, \hat{\mathbf{y}}) \\ & + \lambda_{OOBL} OOBL\left(\hat{\mathbf{Y}}^{\mathbf{M}}\right) \\ & + \lambda_{SL} SL\left(\hat{\mathbf{Y}}, \mathbf{M}\right) \\ & + \lambda_{L2} L2(\boldsymbol{\theta}) . \end{aligned} \tag{7}$$

Here, the lambdas are scalar multipliers used to weigh the four loss terms relative to each other. This study used $\lambda_{MSE} = 1$, $\lambda_{OOBL} = 10^{-3}$, $\lambda_{SL} = 20$, and $\lambda_{L2} = 10^{-3}$. These values were chosen to ensure that MSE was the numerically dominant term while the other terms still made up a non-negligible part of the loss function. In practice, the other terms were between one and two orders of magnitude lower than the MSE. This ensures that predicting an accurate mean fat content remains the primary optimization objective.

To train the modified U-Net model using the loss function in Equation (7), the Adam optimizer [44] was used with an initial learning rate of $10^{-3}$, and hyperparameters $\beta_1 = 0.9$, and $\beta_2 = 0.999$. The best model was chosen based on the validation MSE computed by Equation (3) using the entire validation set as a single batch[2] The model selection is based on MSE instead of TL, as this study's primary interest is a solution minimizing MSE while using the other loss terms to guide the properties of such a solution (i.e., smoothly varying and inside the 0–100 range). However, as evident in Section 3, MSE dominates the other terms in TL in practice. Therefore, using TL instead of MSE for model selection would likely yield similar results.

For a given fold in the five-fold cross-validation, during training, the current best weights (the best weights seen so far) and the associated optimizer state at that point in time were tracked. After each epoch, the model was evaluated on the validation set. The best weights were defined as those having yielded the minimum validation MSE so far.

The learning rate was left unchanged for the first 30 epochs, called the burn-in period. After the burn-in epochs, counters for early stopping and learning rate reduction were initiated. After 10 epochs without reaching a new set of best weights, the learning rate was reduced by a factor of 10 but not below $10^{-7}$. Immediately before lowering the learning rate, the model was restored to its current best weights, and the optimizer's state was restored to its state at that time. If, after 30 epochs, no new set of best weights was achieved, training was halted, and the current best weights were restored and returned. If the early stopping was not triggered, the training was halted after 250 epochs. In practice, all training runs were halted by the early stopping criterion, not the 250 epoch limit.

As for PLS, the aim was to have a U-Net trained on all the available data in the CV set. This could be achieved for PLS by averaging the hyperparameter (number of components) over the CV folds and refit on the entire CV set using this hyperparameter setting. For U-Net, however, averaging hyperparameters such as early stopping epochs and epochs for learning rate reduction provide no guarantee that a U-Net trained with these hyperparameters will have its weights end up in a meaningful place. Instead, an ensemble was constructed from all five U-Nets to have a U-Net trained on the entire CV set. Then, the ensemble's prediction is simply the uniform average of its constituents' predictions. This ensemble U-Net will simultaneously have been trained and validated on all images in the CV set. Constructing such an ensemble has previously yielded good results for predicting protein content in NIR-HSI images of grain [39].

## 3 | Results

### 3.1 | U-Net Convergence

Figure 5 illustrates the total and individual loss terms for CV split 1 throughout the training epochs. It highlights the epochs for which the burn-in period completes (epoch 30), the current best weights (as determined by minimization of validation MSE), and the epochs for learning rate reduction. Graphs for all five CV splits are shown in Appendix B in Figure B1. As the training set is continuously visited in batches, it is common to report a mean value of the evaluated metrics computed over the batches of the training set. Conversely, the same metrics are typically reported for the validation set as computed with the final weights of an epoch. However, the model weights are updated after each batch; therefore, such a comparison between the training and validation metrics is based on different weights of the model. Instead, to enable a direct comparison, an evaluation of the final model after each epoch on the entire training and validation set was performed, thus allowing for a direct comparison between training and validation set metrics.

Inspecting Figure 5, it was clear that there was a downward trend for all loss terms, although it was subject to much noise. The OOBL term quickly converged to zero, indicating that the U-Net had correctly learned to constrain predictions to the 0%–100% range. Additionally, the MSE term was, by far, the dominant term in the loss function. Indeed, this indicated that the optimization process did not need to compromise the precision of fat content predictions to generate smooth chemical maps that did not contain out-of-bounds predictions. The remaining splits, shown in Figure B1, exhibited a similar nature of downward albeit noisy trends for all loss terms.

The graphs' noisy nature was attributed to the relatively small dataset size compared with the rather large U-Net. Additionally, like for Ronneberger et al. [17], a batch size of one was used due to memory constraints, which may also lead to noisy gradients. The weight restoration and learning rate reduction handle this instability. In particular, if the validation loss does not improve, the weights are restored to a known good state, and the optimization continues from that point with a reduced learning rate to minimize fluctuations in the loss. Indeed, Figure 5 shows that the magnitude of fluctuations in the loss decreases as the number of epochs increases, particularly when the learning rate is reduced. The same holds for the other CV splits as shown in Appendix B.

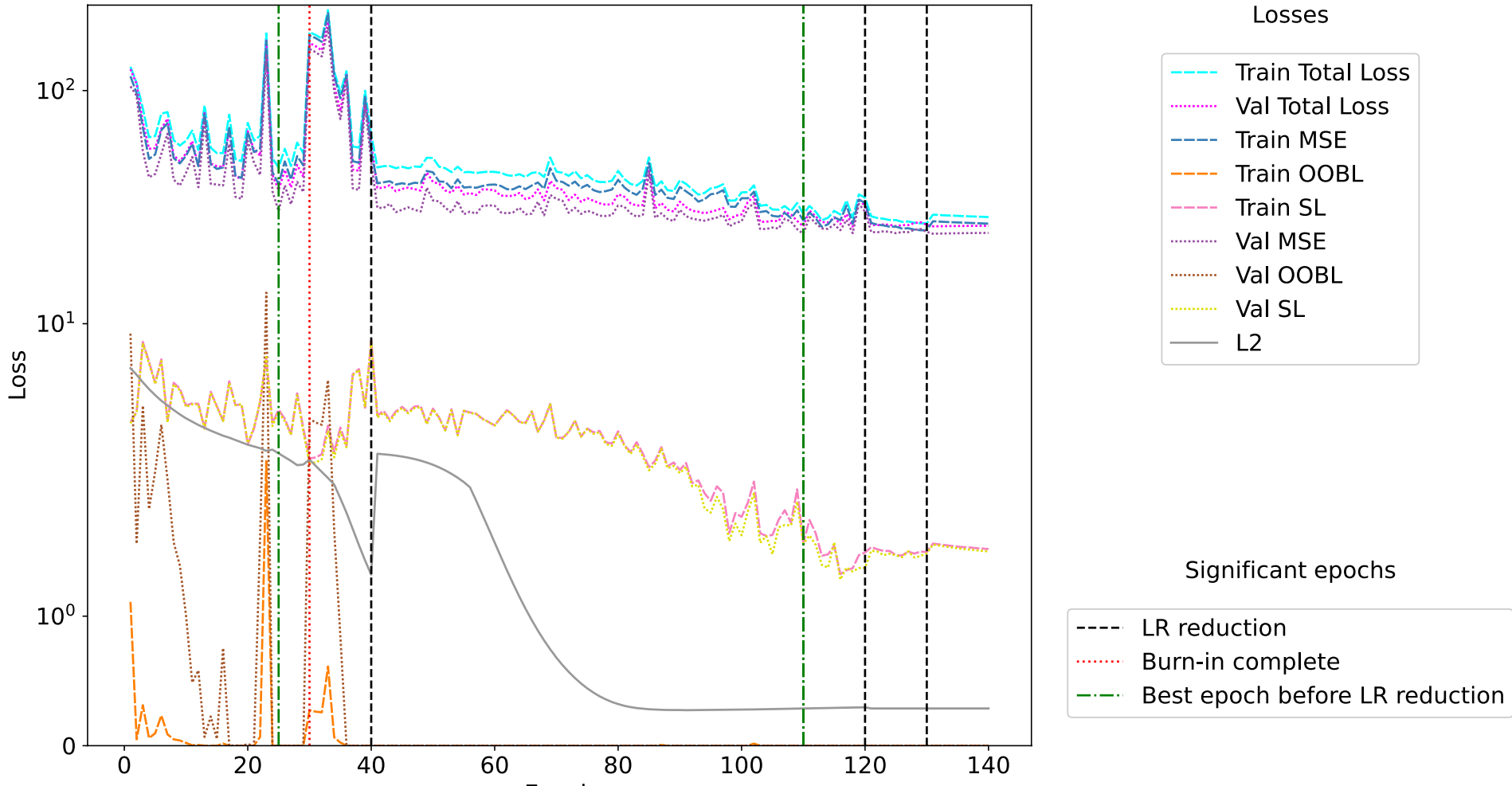

**FIGURE 5** | Evolution of total loss and individual loss terms for CV split 1 (after multiplication with their respective weights).

The code for training and evaluating U-Net is released[3] under the LPGL 3.0 License. The U-Net ensemble's weights are released[4] under CC BY-NC 4.0.

### 3.2 | Pork Belly Mean Fat Predictions

The mean fat content was the primary quality parameter for the predictions. The reference values were given for whole pork bellies, not individual slices. Thus, to get a mean pork belly prediction for U-Net, the pixel-wise predictions for all pixels inside the masks of all slices belonging to a given pork belly were averaged. Conversely, the PLS model could be evaluated directly on the pork belly-wise mean spectra. Figure 6 shows the belly-wise predictions for U-Net and PLS and Table 2 summarizes their test set RMSEs. Appendix C shows additional metrics.

U-Net outperformed PLS on fat classes F2, F4, and F5, while PLS on mean belly spectra showed the best performance on F1, and PLS averaged over chemical maps showed the best performance on F3. Notice also that the U-Net generally had a slightly lower RMSE on the test set than either PLS method. U-Net showed a minimal discrepancy between RMSE on the CV and test sets compared with PLS. In addition to showing RMSE, the line of best fit was computed for each method for both the CV and the test sets. For these lines, sYX was reported, which is the RMSE around the line of best fit but with two fewer degrees of freedom due to the slope and intercept computations. The sYX values are close to the corresponding RMSE values for both U-Net and PLS, indicating well-calibrated models.

### 3.3 | Chemical Maps

In addition to the mean fat predictions, the chemical maps and distributions of the pixel-wise predictions were of high interest to justify using the NIR-HSI technology. Figure 7 shows an example of an input image and the predicted chemical maps by U-Net and PLS, respectively. Figure 8 shows predicted chemical maps for a pork belly from each fat class.

Interestingly, PLS makes predictions that most resemble spatially uncorrelated noise and massive deviations from the 0%–100% range. However, when averaging the otherwise noisy predictions, PLS achieves a mean fat prediction not too far from the reference value as seen in Figure 6c. Every PLS chemical map prediction in the dataset, CV and test sets alike, exhibits this phenomenon of noisy chemical maps with somewhat accurate mean values. Conversely, U-Net generated chemical maps that were spatially structured while constraining itself to the 0%–100% range and simultaneously achieving a mean prediction close to the reference value. Furthermore, while PLS always makes predictions for which the histogram resembles a normal distribution (such as shown in Figure 7e), U-Net predicts a different distribution for each pork belly slice.

Consider the U-Net generated chemical maps for the pork belly in the second row in Figure 8a. A systematic line with uniform fat prediction appears at the very top of these chemical maps. This line is a border effect that arises if, and only if, the pork belly slice is close to the top or bottom of the original image and subsequently padded with the background spectrum. For example, this border effect would arise at the bottom but not at the top of the U-Net generated chemical map for the pork belly slice in Figure 4b. The border effect does not arise on images that are not padded at the top or bottom with the background spectrum. For example, the generated chemical map for the pork belly slice in Figure 4a would not have the border effect.

#### 3.3.1 | Spatial Distribution Analysis

Here, the spatial distribution of pixel-wise predictions is analyzed. First, a qualitative analysis is presented based on the correlation between fat content and firmness of the bellies, measured by the finger pressure method [45]. Then, a statistical analysis of the prediction distribution was conducted.

For the data used in this study, Albano-Gaglio et al. [2] have measured firmness in all 15 regions of the pork bellies (illustrated in Figure 1) and reported a negative correlation with fat content. In general, the dorsal cranial part of a pork belly will have a higher fat content than the ventral caudal part. In Figure 8, this tendency is visible on the chemical maps generated by U-Net. On the other hand, those generated by PLS show no clear structure upon visual inspection.

Figure 9 plots chemical map predictions on the test set, averaged over each of the 15 regions in the pork bellies, against finger pressure for the same regions. Here, a clear negative correlation exists between predicted fat content and finger pressure for both U-Net (-0.91) and PLS (-0.96). Additionally, U-Net and PLS predict higher fat content for the dorsal cranial sections than for the ventral caudal sections.

Inspired by Herrero-Langreo et al. [46], an analysis of the variance of the spatial distribution of the predicted chemical maps was performed. For this, a predicted chemical map, $\hat{\mathbf{Y}}$, was considered with its mask, $\mathbf{M}$, and its mean prediction, $\hat{\mathbf{y}}$ (here, the batch index is omitted for readability). The variance ($\sigma^2$) of the part of $\hat{\mathbf{Y}}_{h,w}$ lying inside $\mathbf{M}$ was analyzed. Quantifying how much of the variance was due to spatial correlation was of particular interest. To this end, the so-called nugget effect ($C_0$) was computed, which is the semi-variogram ($\gamma^2$) evaluated as the pair-wise distance goes to zero [47].

$$\gamma^2\left(\hat{\mathbf{Y}}, \mathbf{M}, \delta_h, \delta_w\right) = \frac{1}{2} \frac{\sum_{h=1}^{H-\delta_h} \sum_{w=1}^{W-\delta_w} \mathbf{M}_{h,w} \left( \left( \hat{\mathbf{Y}}_{h,w} - \hat{\mathbf{Y}}_{h+\delta_h, w+\delta_w} \right)^2 \right)}{\sum_{h=1}^{H-\delta_h} \sum_{w=1}^{W-\delta_w} \mathbf{M}_{h,w}}. \tag{8}$$

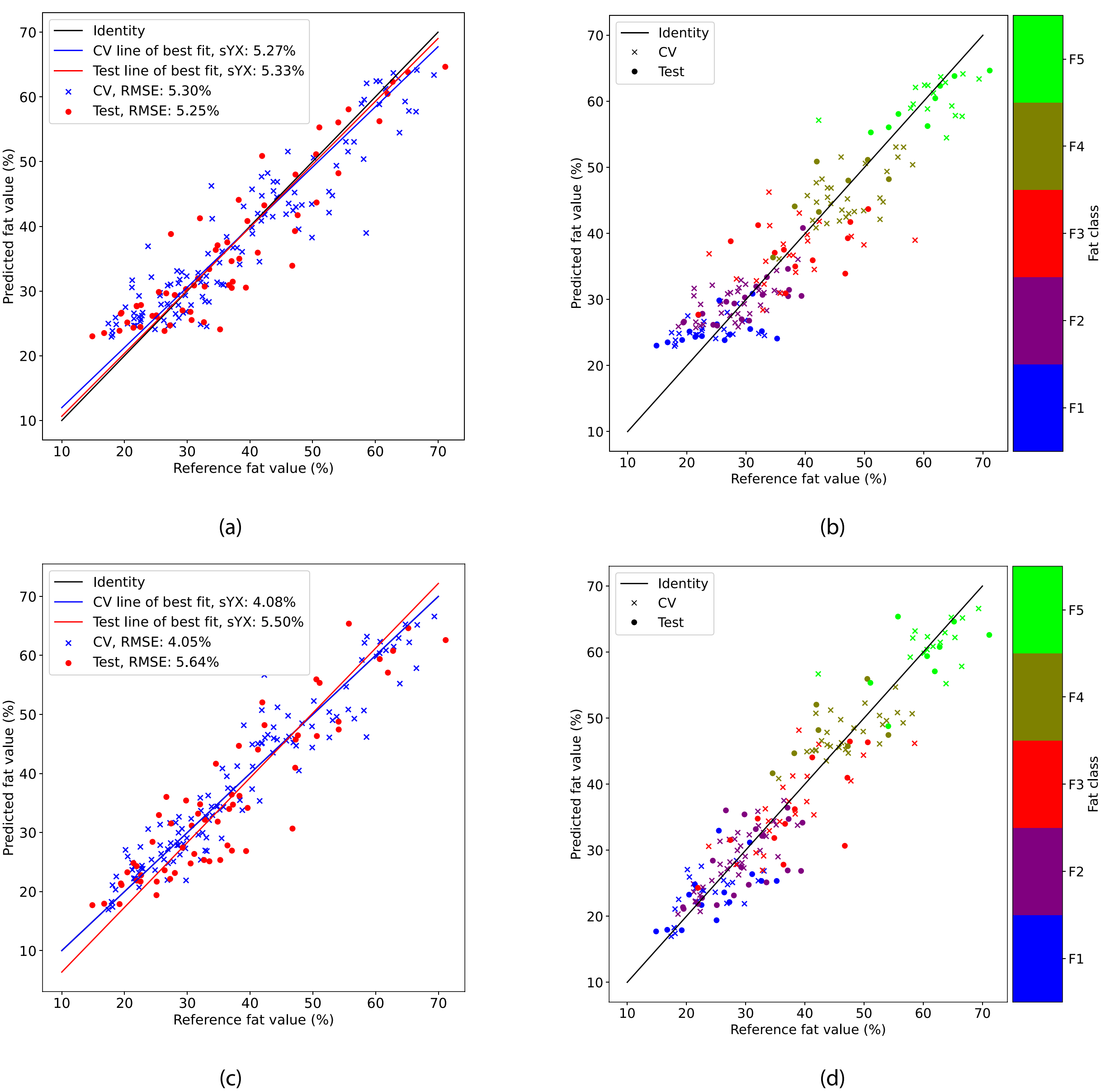

**FIGURE 6** | Pork belly-wise mean fat predictions. (a) U-Net. (b) U-Net with fat classes highlighted. (c) PLS. (d) PLS with fat classes highlighted.

Here, between pairs, $\delta_h$ and $\delta_w$ are the distances in height and width, respectively. As discrete images were used, $C_0$ was computed as the mean of the semi-variogram evaluated at $(\delta_h, \delta_w) = (0, 1)$ and $(\delta_h, \delta_w) = (1, 0)$

Under mild assumptions, subtracting $C_0$ from the total vari-

$$C_0\left(\widehat{\mathbf{Y}}, \mathbf{M}\right) = \frac{\gamma^2\left(\widehat{\mathbf{Y}}, \mathbf{M}, 0, 1\right) + \gamma^2\left(\widehat{\mathbf{Y}}, \mathbf{M}, 1, 0\right)}{2} \quad (9)$$

ance, $\sigma^2$, of $\widehat{\mathbf{Y}}$ inside $\mathbf{M}$ gives the covariance between neighboring pixels [48]. Thus, $C_0$ is a measure of spatially uncorrelated variance. Consequently, $\frac{C_0}{\sigma^2}$ and $1 - \frac{C_0}{\sigma^2}$ are the ratios of spatially uncorrelated, respectively, correlated variance. Table 3 shows the means of these measures computed over the test set for both PLS and U-Net. Almost all of the variance in U-Net–generated chemical maps was spatially correlated. In contrast, only around three percent of the variance in PLS-generated chemical maps was spatially correlated. The low $C_0$ value for U-Net predictions can likely be attributed to the following equivalence between $C_0$ Equation (9) and U-Net's smoothness loss (SL) term Equation (5): $C_0(\widehat{\mathbf{Y}}, \mathbf{M}) = \frac{1}{4} SL\left(\widehat{\mathbf{Y}}, \mathbf{M}\right)$. As such, U-Net was directly optimized to minimize $C_0$, which was not the case for PLS.

Considering all the results presented in this section, a reader might wonder if a smoothened version of the PLS predictions would show the same underlying structure revealed in the U-Net predictions. Indeed, this was attempted. Smoothing the PLS predictions to a sufficiently high degree keeps the pixel-wise predictions inside the 0%–100% range. However, to get a low $C_0$, the PLS-generated chemical maps must be smoothened to such a large scale that any fine spatial detail vanishes. Thus, smoothing the PLS predictions does not yield chemical maps equivalent to those generated by U-Net.

**TABLE 2** | RMSE (in %) of pork belly-wise fat predictions on the test set for U-Net and PLS.

| | Fat class | | | | | |
|---|---|---|---|---|---|---|
| **Model** | **F1** | **F2** | **F3** | **F4** | **F5** | **All** |
| U-Net | 5.40 | 4.36 | 7.29 | 4.69 | 3.41 | 5.25 |
| PLS | 4.84 | 5.54 | 6.12 | 6.59 | 5.51 | 5.64 |
| Reference std. | 5.88 | 6.35 | 8.38 | 6.39 | 6.08 | 13.2 |
| Reference mean | 25.0 | 29.8 | 38.4 | 44.1 | 60.3 | 36.1 |

*Note:* The reference means and standard deviations are shown for interpretation of the models' RMSE values.

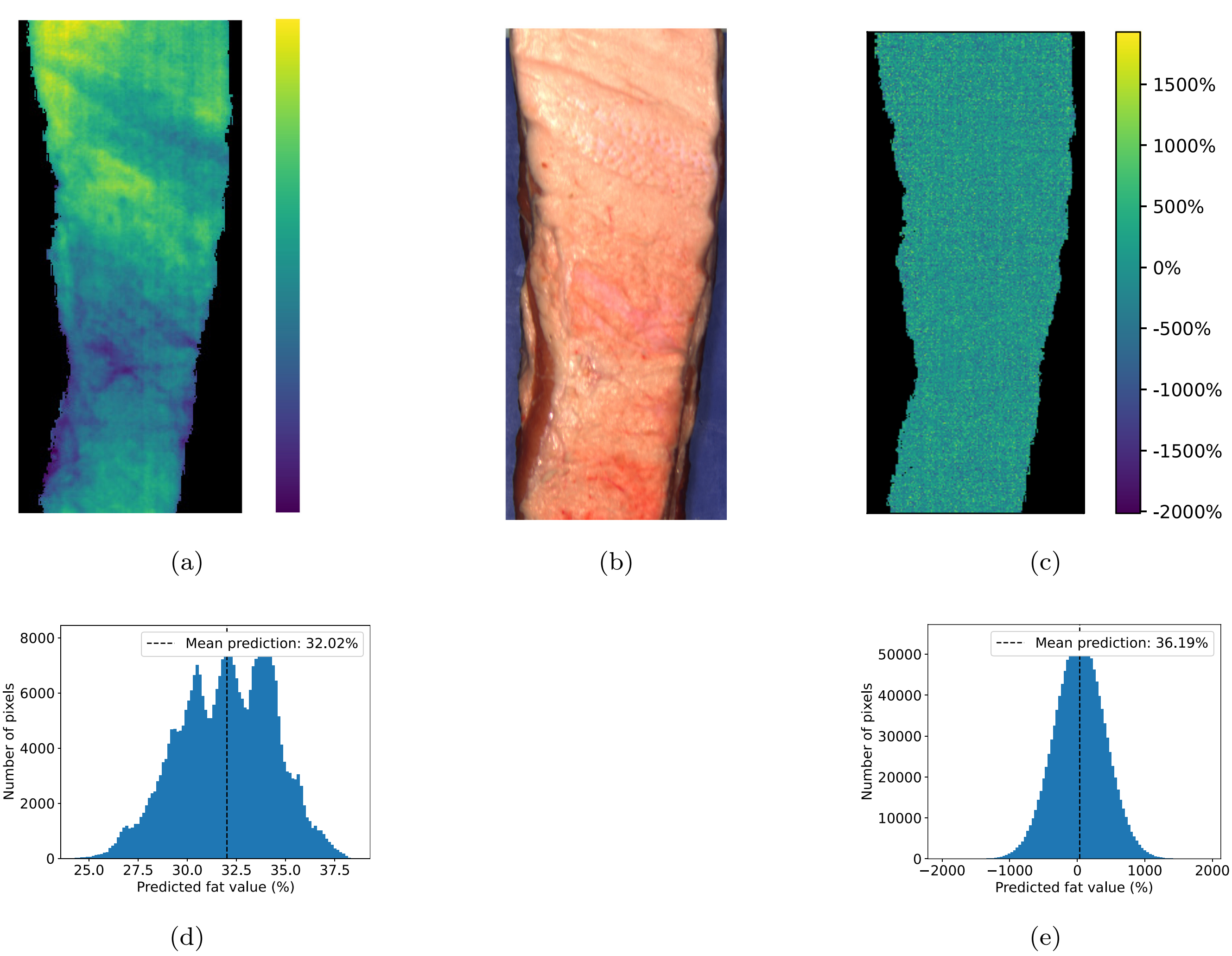

**FIGURE 7** | Example chemical map predictions on a slice of pork belly from the test set. (a) Masked U-Net prediction. (b) Input image with a reference mean fat content of 32.79%. For U-Net, the input was the padded version shown in Figure 4a. (c) Masked PLS prediction. (d) Histogram for the pixel-wise U-Net predictions. (e) Histogram for the pixel-wise PLS predictions.

## 4 | Discussion

### 4.1 | NIR-HSI Depth Penetration for Fat Content Analysis

This study's modeling process was based on the assumption that a correlation exists between the information in the hyperspectral images and the reference fat values. The reference fat values were measured chemically using the entire bellies, while the NIR-HSI camera captures only light reflected on the surface and slightly below the surface due to subsurface scattering. As such, it is unlikely that the models can predict total fat without seeing the entire bellies in depth. However, as shown in Figure 9, the chemical maps generated by U-Net and PLS exhibit a high negative correlation with finger pressure. Finger pressure, in turn, has a high negative correlation with fat content [2], which indicates that the NIR-HSI predictions contain relevant information about the fat distribution in the belly. Thus, it was hypothesized that the models had learned to predict total fat from a measure of subcutaneous fat. It is emphasized, however, that the methods can readily be applied to other image modalities, such as X-ray images and CT scans, which allow image analysis of the entire bellies in depth.

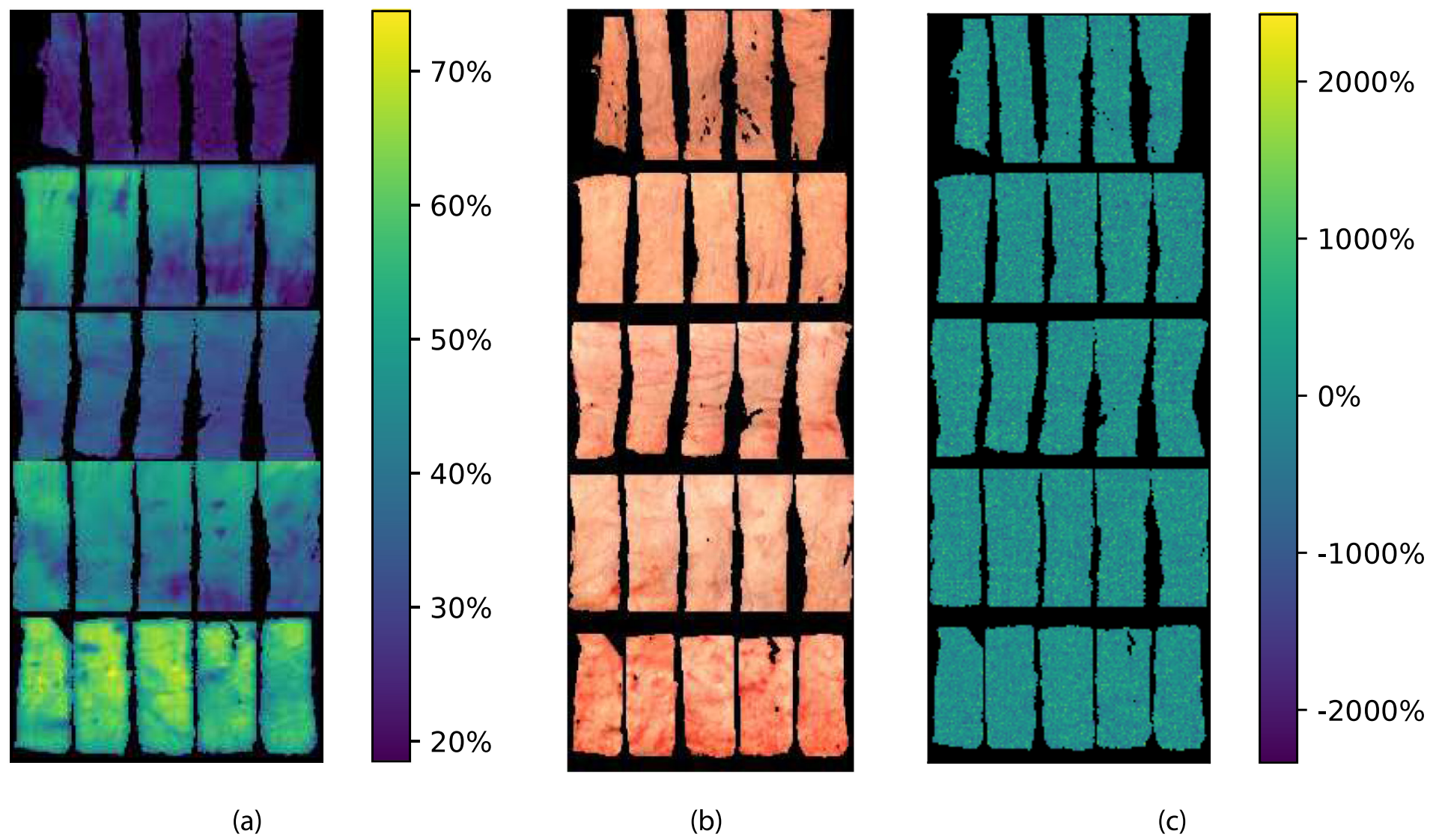

**FIGURE 8** | U-Net and PLS generated chemical maps for a test set pork belly from each fat class, F1–F5. Each row represents an entire pork belly. The first row belongs to fat class F1, increasing toward the last row belonging to F5. As in Figure 1, the five slices of each belly are ordered from left (cranial) to right (caudal), and each slice is oriented with the dorsal side up and the ventral side down. (a) U-Net predictions. (b) RGB illustrations of pork bellies. (c) PLS predictions.

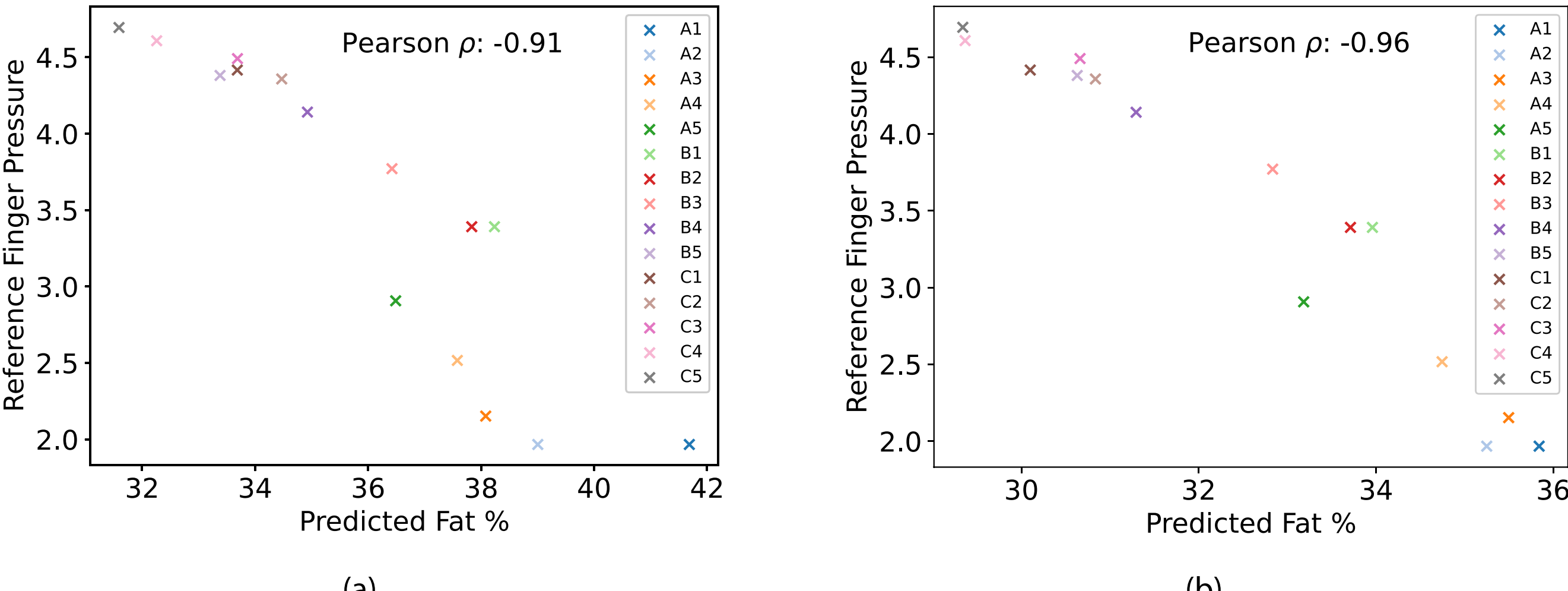

**FIGURE 9** | Relationship between mean predicted fat percentages, (a) U-Net, (b) PLS, and mean finger pressure for each of the 15 sections of pork bellies in the test set (see regions in Figure 1). The Pearson correlation coefficients indicate a high negative linear correlation for both models.

## 4.2 | Bias and Scale Correction

Consider the U-Net and PLS predictions in Figure 6. For PLS, the line of best fit for the CV set was guaranteed to be the identity due to the least squares nature of PLS. However, this was not the case for U-Net. Deep learning regressors do not guarantee to find a least squares solution [49], especially when training with bulk references and making sample-wise predictions [50]. However, the U-Net's CV set sYX was very close to its RMSE. Furthermore, bias and scale correction do not change the visual appearance of the chemical maps, nor do they change the values of the ratios in Table 3. For these reasons, no bias or scale corrections were performed.

## 4.3 | Chemical Maps

PLS and the modified U-Net had approximately the same performance when evaluated on the mean fat predictions for pork bellies (Figure 6). However, as evident by the spatial statistics in Table 3, U-Net generated much more meaningful chemical maps than PLS. In addition to generating spatially aware, locally

**TABLE 3** | Total variance, $\sigma^2$, and spatially uncorrelated variance, $C_0$.

| Model | Metric | | | |
|---|---|---|---|---|
| | $\sigma^2$ | $C_0$ | $\frac{C_0}{\sigma^2}$ | $1-\frac{C_0}{\sigma^2}$ |
| PLS | 130276 | 127249 | 0.98 | 0.02 |
| U-Net | 25.17 | 0.01 | 0.001 | 0.999 |

*Note:* As PLS and U-Net predictions have vastly different variances, the ratios of spatially uncorrelated, respectively, correlated variance are better suited for comparison. The values shown are means computed over the test set predictions.

smooth chemical maps, U-Net almost always stayed inside the 0%–100% range. This was unlike PLS, where most predictions lay outside this range.

The U-Net's loss function does not have a term to accommodate the higher fat concentration in the dorsal cranial section than in the ventral caudal section. Still, this exact behavior emerged when analyzing the distributions of the predicted chemical maps. Thus, the proposed learning framework could model such correlations without explicitly being optimized against them, indicating that the U-Net had learned to generate meaningful chemical maps. Across all five fat classes, U-Net and PLS both predicted higher fat concentrations in the dorsal cranial sections and lower fat concentrations in the ventral caudal sections. U-Net, however, predicted chemical maps with much finer detail than PLS and thus, based on the fat distributions, allowed for a more precise division of pork bellies for downstream processing and value generation.

PLS chemical map predictions were accurate on average and highly correlated with finger pressure (Figure 9) in the 15 subsections of the pork belly. As such, it may seem that simply applying a smoothing operation to the PLS chemical map predictions would reveal the same spatial structure as those exhibited in U-Net–generated chemical maps. Such smoothing was attempted but to no avail. To reveal spatial structure, the PLS-generated chemical maps had to be smoothened to such a degree that no fine detail remained.

Due to the lack of pixel-wise reference values, it was not possible to directly verify the correctness of predicted chemical maps with this dataset alone. However, the low RMSE shown in Figure 6b, the spatial structure shown in Figure 8a, and the spatial correlations shown in Figure 9a and Table 3 all indicated that the U-Net had learned to generate meaningful chemical maps to the extent of what could be assessed with this dataset.

### 4.4 | U-Net Extensions

In the U-Net's loss function, Equation (7), multipliers were used to weigh the different terms relative to each other. The values taken by these multipliers were chosen manually during initial experimentation. The values were chosen such that the numerically dominant term in the loss function would be the mean squared error, while the other parts still made up a small but significant part. The basis for this choice was the main interest in the least squares solution. At the same time, the other terms sought to guide the properties of this solution such that it exhibited local smoothness and contained only values inside the 0%–100% range.

Instead of the padding scheme used in this study, zero-padding could have been applied to either the input hyperspectral images or the convolution and pooling operations. These zero-padding schemes were attempted during initial experimentation but yielded worse border effects than those arising with the chosen scheme.

While the goal of this study was to serve as a proof of concept, it was likely possible to further decrease the mean squared error of the U-Net by performing more elaborate studies regarding the values of hyperparameters, including the previously mentioned loss-multipliers, learning rate, optimizer, and all other choices that go into training and evaluating a deep learning model. While such performance optimization was not of interest in this study, it is possible to search for more optimal hyperparameters using, for example, grid-search or a Bayesian approach such as a Tree-Structured Parzen Estimator (TPE) [51] as implemented by Akiba et al. [52].

In this study, the fat content of pork bellies was analyzed. The dataset includes references for additional parameters [8]. Future developments of this work involve generating chemical maps for multiple parameters using a single model, e.g., a U-Net. A natural extension of the presented loss function would be suitable in this case. First, each parameter should be scaled by its standard deviation to ensure equal contribution to the total loss. Afterward, it is simple to compute an MSE term, Equation (3), and an SL term, Equation (5), for each parameter. Interestingly, optimizing the model to regress multiple parameters yields additional possibilities related to the out-of-bounds loss, Equation (4), that aims to constrain the pixel-wise prediction of each parameter to values between 0 and 100 percent. In addition to constraining each parameter, the sum of parameters should also be constrained in this range. If all parameters can be measured, they should sum to 100 percent. Thus, it is hypothesized that the generalization of the presented work to multiple parameters is not only feasible but perhaps even beneficial compared with the single-parameter case.

Another possible direction for extending this work can be to maintain U-Net's chemical map generation capability while also having it learn to perform segmentation between foreground (pork belly) and background, a task for which it was initially designed [17].

## 5 | Conclusion

This study has proposed a novel end-to-end deep learning approach based on U-Net for chemical map generation using hyperspectral images of pork bellies with associated mean fat reference values as a case study. By training a modified U-Net with a robust optimization scheme and a multi-faceted customized loss function, using only a mean reference value, the model has learned to generate chemical maps with a high degree of spatial correlation and to make predictions strictly inside the 0%–100% range of fat values. These findings contrast the current PLS-based pixel-wise approach to chemical map generation that will generate chemical maps with a low degree of spatial correlation and predictions ranging from −2000% to 2000%, significantly exceeding the realm of the physically possible. Simultaneously, the U-Net–based approach achieves a lower RMSE than PLS when averaging the chemical maps for comparison with the mean reference values. Thus, this study indicates that the U-Net–based approach enables the generation of chemical maps with meaningful spatial features, emphasizing that the analysis of hyperspectral images can benefit from joint consideration of the spectral and spatial features.

Having shown that deep learning segmentation models such as U-Net can be trained to generate chemical maps from input images directly, further developments in this area are expected to show that they can simultaneously generate chemical maps for multiple parameters and a segmentation mask to distinguish between background and the product of interest in a single forward pass.

---

### Acknowledgments

We want to thank the IRTA technicians Albert Brun, Agustí Quintana, Albert Rossell, Adrià Pacreu, Cristina Canals, and Joel González, for their help in collecting the data used in this project. We also want to thank José M. Martínez for his contribution to the analysis of fatty acids. The CERCA program from the Generalitat de Catalunya is also acknowledged. We also thank Dr. Aneesh Chauhan from WUR for his advice on using complete HSI images for model training.

### Endnotes

[1] The square root was omitted for faster computation. As the square root is a monotonically increasing function, omitting it does not change the model's optimum.

[2] The whole validation set could not be stored on the GPU. Therefore, when encountering a sample, the squared errors were accumulated, returning the mean when all samples had been seen. This computation is mathematically equivalent to using the entire validation set as a batch.

[3] https://github.com/sm00thix/PorkBellyHSI.

[4] https://huggingface.co/Sm00thix/unet_chemical_map.

### References

1. M. Mäkelä, P. Geladi, M. Rissanen, L. Rautkari, and O. Dahl, "Hyperspectral Near Infrared Image Calibration and Regression," *Analytica chimica acta* 1105 (2020): 56–63, https://doi.org/10.1016/j.aca.2020.01.019.

2. M. Albano-Gaglio, C. Zomeño, J. F. Tejeda, et al., "Pork Belly Quality Variation and Its Association With Fatness Level," *Meat Science* 213 (2024): 109482, https://doi.org/10.1016/j.meatsci.2024.109482.

3. F. Tao, H. Yao, Z. Hruska, et al., "Raman Hyperspectral Imaging as a Potential Tool for Rapid and Nondestructive Identification of Aflatoxin Contamination in Corn Kernels," *Journal of Food Protection* 87 (2024): 100335, https://doi.org/10.1016/j.jfp.2024.100335.

4. R. E. Boseley, N. J. Sylvain, L. Peeling, M. E. Kelly, and M. J. Pushie, "A Review of Concepts and Methods for FTIR Imaging of Biomarker Changes in the Post-Stroke Brain," *Biochimica et Biophysica Acta (BBA) - Biomembranes* 2024 (1866): 184287, https://doi.org/10.1016/j.bbamem.2024.184287.

5. J. M. Amigo, H. Babamoradi, and S. Elcoroaristizabal, "Hyperspectral image analysis. A Tutorial," *Analytica Chimica Acta* 896 (2015): 34–51, https://doi.org/10.1016/j.aca.2015.09.030.

6. H. Wold, "Estimation of Principal Components and Related Models by Iterative Least Squares," *Multivariate Analysis* 1 (1966): 391–420.

7. S. Wold, M. Sjöström, and L. Eriksson, "PLS-Regression: A Basic Tool of Chemometrics," *Chemometrics and Intelligent Laboratory Systems* 58 (2001): 109–130, https://doi.org/10.1016/S0169-7439(01)00155-1.

8. M. Albano-Gaglio, P. Mishra, S. W. Erasmus, et al., "Visible and Near-Infrared Spectral Imaging Combined With Robust Regression for Predicting Firmness, Fatness, and Compositional Properties of Fresh Pork Bellies," *Meat Science* 219 (2025): 109645, https://doi.org/10.1016/j.meatsci.2024.109645.

9. G. ElMasry, N. Mandour, Y. Ejeez, et al., "Multichannel Imaging for Monitoring Chemical Composition and Germination Capacity of Cowpea (*Vigna unguiculata*) Seeds During Development and Maturation," *Crop Journal* 10 (2022): 1399–1411, https://doi.org/10.1016/j.cj.2021.04.010.

10. G. M. ElMasry, E. Fulladosa, J. Comaposada, S. S. Al-Rejaie, and P. Gou, "Selection of Representative Hyperspectral Data and Image Pretreatment for Model Development in Heterogeneous Samples: A Case Study in Sliced Dry-Cured Ham," *Biosystems Engineering* 201 (2021): 67–82, https://doi.org/10.1016/j.biosystemseng.2020.11.008.

11. A. Thumm, M. Riddell, B. Nanayakkara, J. Harrington, and R. Meder, "Near Infrared Hyperspectral Imaging Applied to Mapping Chemical Composition in Wood Samples," *Journal of Near Infrared Spectroscopy* 18 (2010): 507–515, https://doi.org/10.1255/jnirs.909.

12. F. Jamme and L. Duponchel, "Neighbouring Pixel Data Augmentation: A Simple Way to Fuse Spectral and Spatial Information for Hyperspectral Imaging Data Analysis," *Journal of Chemometrics* 31 (2017): e2882, https://doi.org/10.1002/cem.2882.

13. J. L. Xu and A. Gowen, "Spatial-Spectral Analysis Method Using Texture Features Combined With PCA for Information Extraction in Hyperspectral Images," *Journal of Chemometrics* 34 (2020): e3132, https://doi.org/10.1002/cem.3132.

14. B. Gaci, F. Abdelghafour, M. Ryckewaert, et al., "A Novel Approach to Combine Spatial and Spectral Information From Hyperspectral Images," *Chemometrics and Intelligent Laboratory Systems* 240 (2023): 104897, https://doi.org/10.1016/j.chemolab.2023.104897.

15. N. Gorretta, G. Rabatel, C. Fiorio, C. Lelong, and J. M. Roger, "An Iterative Hyperspectral Image Segmentation Method Using a Cross Analysis of Spectral and Spatial Information," *Chemometrics and Intelligent Laboratory Systems* 117 (2012): 213–223, https://doi.org/10.1016/j.chemolab.2012.05.004.

16. P. Mishra, M. Albano-Gaglio, and M. Font-i Furnols, "A Short Note on Deep Contextual Spatial and Spectral Information Fusion for Hyperspectral Image Processing: Case of Pork Belly Properties Prediction," *Journal of Chemometrics* 38 (2024): e3552, https://doi.org/10.1002/cem.3552.

17. O. Ronneberger, P. Fischer, and T. Brox, "U-Net: Convolutional Networks for Biomedical Image Segmentation," in *Medical Image Computing and Computer-Assisted Intervention–MICCAI 2015: 18th International Conference, Munich, Germany, October 5-9, 2015, Proceedings*, part III 18 (Springer, 2015): 234–241, https://doi.org/10.1007/978-3-319-24574-4.

18. R. Archana and P. S. E. Jeevaraj, "Deep Learning Models for Digital Image Processing: A Review," *Artificial Intelligence Review* 57 (2024): 11, https://doi.org/10.1007/s10462-023-10631-z.

19. R. Azad, E. K. Aghdam, A. Rauland, et al., "Medical Image Segmentation Review: The Success of U-Net," *IEEE Transactions on Pattern Analysis and Machine Intelligence* 46, no. 12 (2024): 10076–10095, https://doi.org/10.1109/TPAMI.2024.3435571.

20. Z. Luo, W. Yang, Y. Yuan, R. Gou, and X. Li, “Semantic Segmentation of Agricultural Images: A Survey,” *Information Processing in Agriculture* 11 (2024): 172–186, https://doi.org/10.1016/j.inpa.2023.02.001.

21. N. Siddique, S. Paheding, C. P. Elkin, and V. Devabhaktuni, “U-Net and Its Variants for Medical Image Segmentation: A Review of Theory and Applications,” *IEEE Access* 9 (2021): 82031–82057, https://doi.org/10.1109/ACCESS.2021.3086020.

22. M. Barker and W. Rayens, “Partial Least Squares for Discrimination,” *Journal of Chemometrics: A Journal of the Chemometrics Society* 17 (2003): 166–173, https://doi.org/10.1002/cem.785.

23. R. D. Snee, “Validation of Regression Models: Methods and Examples,” *Technometrics* 19 (1977): 415–428, https://doi.org/10.1080/00401706.1977.10489581.

24. P. C. Mahalanobis, “On the Generalized Distance in Statistics. Sankhyā,” *Indian Journal of Statistics, Series A (2008-)* 80 (2018): S1–S7, https://www.jstor.org/stable/48723335.

25. H. Hotelling, “Analysis of a Complex of Statistical Variables Into Principal Components,” *Journal of Educational Psychology* 24 (1933): 417, https://doi.org/10.1037/h0071325.

26. K. Pearson, “Liii. on Lines and Planes of Closest Fit to Systems of Points in Space,” *London, Edinburgh, and Dublin Philosophical Magazine and Journal of Science* 2 (1901): 559–572, https://doi.org/10.1080/14786440109462720.

27. R. J. Barnes, M. S. Dhanoa, and S. J. Lister, “Standard Normal Variate Transformation and De-Trending of Near-Infrared Diffuse Reflectance Spectra,” *Applied Spectroscopy* 43 (1989): 772–777, https://doi.org/10.1366/0003702894202201.

28. A. Savitzky and M. J. E. Golay, “Smoothing and Differentiation of Data by Simplified Least Squares Procedures,” *Analytical Chemistry* 36 (1964): 1627–1639, https://doi.org/10.1021/ac60214a047.

29. J. Steinier, Y. Termonia, and J. Deltour, “Smoothing and Differentiation of Data by Simplified Least Square Procedure,” *Analytical Chemistry* 44 (1972): 1906–1909, https://doi.org/10.1021/ac60319a045.

30. Å. Rinnan, F. van den Berg, and S. B. Engelsen, “Review of the Most Common Pre-Processing Techniques for Near-Infrared Spectra,” *TrAC Trends in Analytical Chemistry* 28 (2009): 1201–1222, https://doi.org/10.1016/j.trac.2009.07.007.

31. K. M. Sørensen, F. van den Berg, and S. B. Engelsen, “NIR Data Exploration and Regression by Chemometricsa Primer,” *Near-Infrared Spectroscopy: Theory, Spectral Analysis, Instrumentation, and Applications* 1 (2021): 127–189, https://doi.org/10.1007/978-981-15-8648-4_7.

32. A. Alin, “Comparison of PLS Algorithms When Number of Objects Is Much Larger Than Number of Variables,” *Statistical Papers* 50 (2009): 711–720, https://doi.org/10.1007/s00362-009-0251-7.

33. M. Andersson, “A Comparison of Nine PLS1 Algorithms,” *Journal of Chemometrics* 23 (2009): 518–529, https://doi.org/10.1002/cem.1248.

34. B. S. Dayal and J. F. MacGregor, “Improved PLS Algorithms,” *Journal of Chemometrics* 11 (1997): 73–85, https://www.scirp.org/reference/referencespapers?referenceid=1601772.

35. O. C. G. Engstrøm, E. S. Dreier, B. M. Jespersen, and K. S. Pedersen, “IKPLS: Improved Kernel Partial Least Squares and Fast Cross-Validation Algorithms for Python With CPU and GPU Implementations Using Numpy and Jax,” *Journal of Open Source Software* 9 (2024): 6533, https://doi.org/10.21105/joss.06533.

36. O. C. G. Engstrøm and M. H. Jensen, “Fast Partition-Based Cross-Validation With Centering and Scaling for **X** TX and X^TY,” *Journal of Chemometrics* 39 (2025): e70008, https://doi.org/10.1002/cem.70008.

37. A. Odena, V. Dumoulin, and C. Olah, “Deconvolution and Checkerboard Artifacts,” *Distill* 1 (2016): e3, https://doi.org/10.23915/distill.00003.

38. O. C. G. Engstrøm, E. S. Dreier, K. S. Pedersen, 2021. “Predicting Protein Content in Grain Using Hyperspectral Deep Learning”, in: Proceedings of the IEEE/CVF International Conference on Computer Vision, pp. 1372–1380, https://doi.org/10.1109/ICCVW54120.2021.00158.

39. O. C. G. Engstrøm, E. S. Dreier, B. M. Jespersen, K. S. Pedersen, 2023b. “Improving Deep Learning on Hyperspectral Images of Grain by Incorporating Domain Knowledge From Chemometrics”, in: Proceedings of the IEEE/CVF International Conference on Computer Vision, pp. 485–494, https://doi.org/10.1109/ICCVW60793.2023.00055.

40. K. He, X. Zhang, S. Ren, J. Sun, 2015. “Delving Deep Into Rectifiers: Surpassing Human-Level Performance on ImageNet Classification”, in: Proceedings of the IEEE international conference on computer vision, pp. 1026–1034, https://doi.org/10.1109/ICCV.2015.123.

41. A. Vedaldi, Y. Jia, E. Shelhamer, et al., *Convolutional Architecture for Fast Feature Embedding* (Cornell University, 2014).

42. J. Ansel, E. Yang, H. He, N. Gimelshein, A. Jain, M. Voznesensky, B. Bao, P. Bell, D. Berard, E. Burovski, et al., 2024. Pytorch 2: Faster Machine Learning Through Dynamic Python Bytecode Transformation and Graph Compilation, in: Proceedings of the 29th ACM International Conference on Architectural Support for Programming Languages and Operating Systems, Volume 2, pp. 929–947, https://doi.org/10.1145/3620665.3640366.

43. A. Signoroni, M. Savardi, A. Baronio, and S. Benini, “Deep Learning Meets Hyperspectral Image Analysis: A Multidisciplinary Review,” *Journal of Imaging* 5 (2019): 52, https://doi.org/10.3390/jimaging5050052.

44. D. P. Kingma, J. Ba, 2017. “Adam: A Method for Stochastic Optimization”. https://doi.org/10.48550/arXiv.1412.6980, http://arxiv.org/abs/1412.6980arXiv:1412.6980.

45. S. J. Maw, V. R. Fowler, M. Hamilton, and A. M. Petchey, “Physical Characteristics of Pig Fat and Their Relation to Fatty Acid Composition,” *Meat Science* 63 (2003): 185–190, https://doi.org/10.1016/S0309-1740(02)00069-4.

46. A. Herrero-Langreo, N. Gorretta, B. Tisseyre, et al., “Using Spatial Information for Evaluating the Quality of Prediction Maps From Hyperspectral Images: A Geostatistical Approach,” *Analytica Chimica Acta* 1077 (2019): 116–128, https://doi.org/10.1016/j.aca.2019.05.067.

47. G. Matheron, “Principles of Geostatistics,” *Economic geology* 58 (1963): 1246–1266, https://doi.org/10.2113/gsecongeo.58.8.1246.

48. R. J. Barnes, “The Variogram Sill and the Sample Variance,” *Mathematical Geology* 23 (1991): 673–678.

49. C. Igel and S. Oehmcke, “Remember to Correct the Bias When Using Deep Learning for Regression!,” *KI-Künstliche Intelligenz* 37 (2023): 33–40, https://doi.org/10.1007/s13218-023-00801-0.

50. O. C. G. Engstrøm, E. S. Dreier, B. M. Jespersen, K. S. Pedersen, 2023a. “Analyzing Near-Infrared Hyperspectral Imaging for Protein Content Regression and Grain Variety Classification Using Bulk References and Varying Grain-to-Background Ratios”. https://doi.org/10.48550/arXiv.2311.04042, http://arxiv.org/abs/2311.04042arXiv:2311.04042.

51. J. Bergstra, R. Bardenet, Y. Bengio, and B. Kégl, “Algorithms for Hyper-Parameter Optimization,” *Advances in Neural Information Processing Systems* 24 (2011): 2546–2554.

52. T. Akiba, S. Sano, T. Yanase, T. Ohta, M. Koyama, 2019. “Optuna: A Next-Generation Hyperparameter Optimization Framework”, in: Proceedings of the 25th ACM SIGKDD international conference on knowledge discovery & data mining, pp. 2623–2631, https://doi.org/10.1145/3292500.3330701.

## Appendix A: Modified U-Net Architecture

This appendix shows architecture of the modified U-Net. It consists of the sequentially linked 3D convolution module and the U-Net module. In addition to showing input and output sizes, it shows the sizes of intermediate tensors.

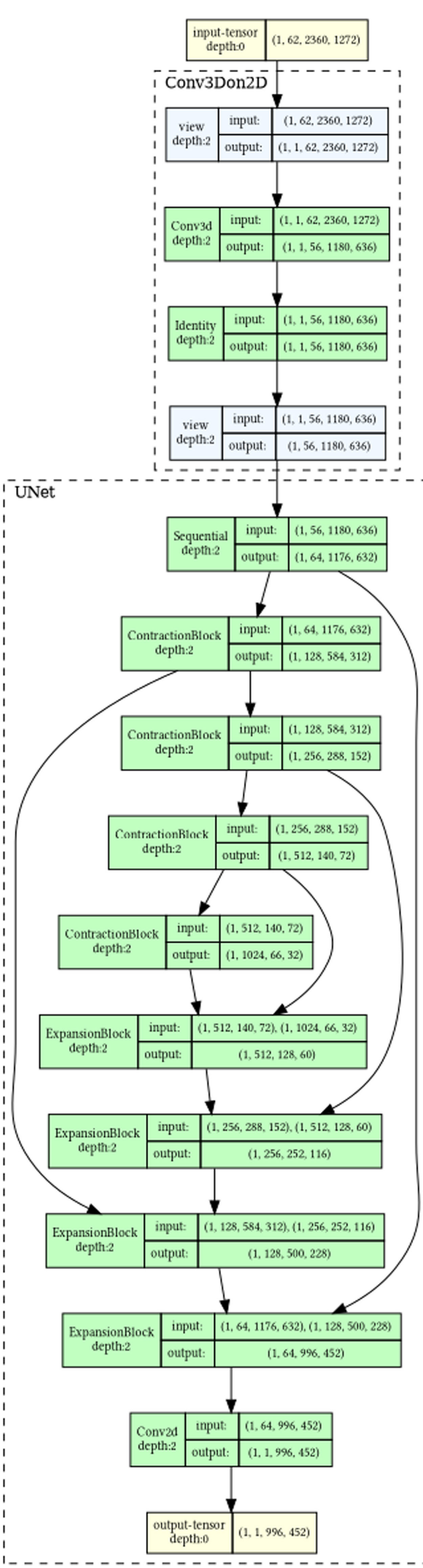

## Appendix B: Loss Evolution for U-Net

Figure B1 shows the loss evolution of U-Net for all five CV splits.

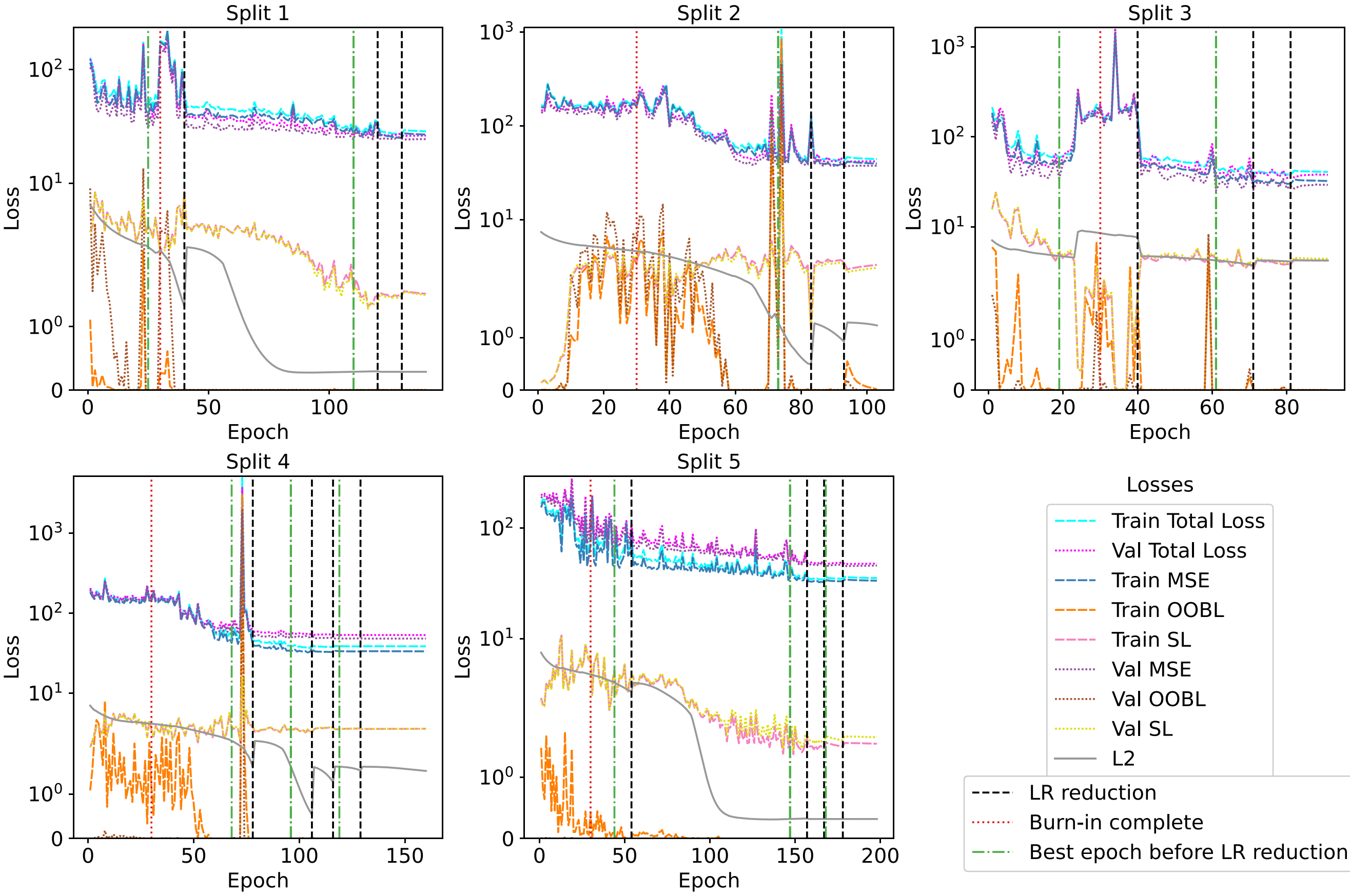

**FIGURE B1** | Evolution of total loss and individual loss terms for all CV splits (after multiplication with their respective weights).

## Appendix C: Metrics

Tables C1 and C2 show $Q^2$ for the test set and $R^2$, RMSE, bias and scale for the CV set and test set. Note that the CV set is everything except the test set and would typically be called the calibration set in chemometric lingo. The bias and scale are the solution to an ordinary least squares regression from the predicted values, $\hat{\mathbf{y}}$ to the reference values, $\mathbf{y}$. They are the values used to compute the lines of best fit in Figure 6.

**TABLE C1** | Common chemometric metrics for the models evaluated on the entire CV set (this is everything except the test set and would be called the calibration set in chemometric lingo).

| | Metric | | | |
|---|---|---|---|---|
| **Model** | $R^2$ | **RMSE** | **bias** | **scale** |
| U-Net | 0.85 | 5.30 | −2.90 | 1.07 |
| PLS | 0.91 | 4.05 | 0.00 | 1.00 |

**TABLE C2** | Common chemometric metrics for the models evaluated on the test set.

| | Metric | | | | |
|---|---|---|---|---|---|
| **Model** | $Q^2$ | $R^2$ | **RMSE** | **bias** | **scale** |
| U-Net | 0.84 | 0.84 | 5.25 | −0.95 | 1.03 |
| PLS | 0.81 | 0.81 | 5.64 | 4.20 | 0.91 |